\documentclass{article}

      \usepackage[nonatbib, preprint]{icbinb_neurips_2020}

\usepackage[utf8]{inputenc} 
\usepackage[T1]{fontenc}    
\usepackage{hyperref}       
\usepackage{url}            
\usepackage{booktabs}       
\usepackage{amsfonts}       
\usepackage{nicefrac}       
\usepackage{microtype}      

\usepackage{amsmath, amsfonts, mathtools}
\usepackage{siunitx}
\usepackage[ruled,vlined]{algorithm2e}

\usepackage{booktabs}
\usepackage{epsfig}
\usepackage{etoolbox}
\usepackage{float}

\usepackage{array}
\newcolumntype{C}[1]{>{\centering\arraybackslash}p{#1}}

\usepackage{caption}
\usepackage{subcaption}
\def\E{{\rm E}}

\def\ELBO{{\rm ELBO}}
\def\KL{\mathbb{D}_{\rm KL} }

\usepackage{xcolor}

\usepackage{multirow}

\title{Semi-supervised Learning by Latent Space Energy-Based Model of Symbol-Vector Coupling}

\author{
Bo Pang$^1$ \hskip1em Erik Nijkamp$^1$  \hskip1em Jiali Cui$^2$ \hskip1em Tian Han$^2$ \hskip1em Ying Nian Wu$^1$ \\
$^1$University of California, Los Angeles \hskip1em $^2$Stevens Institute of Technology
\\ {\small \texttt{\{bopang, enijkamp\}@ucla.edu}} \hskip1em {\small \texttt{\{jcui7, than6\}@stevens.edu}} \hskip1em {\small \texttt{ywu@stat.ucla.edu}}
}

\begin{document}

\maketitle

\begin{abstract}
This paper proposes a latent space energy-based prior model for semi-supervised learning. The model stands on a generator network that maps a latent vector to the observed example. The energy term of the prior model couples the latent vector and a symbolic one-hot vector, so that classification can be based on the latent vector inferred from the observed example. In our learning method, the symbol-vector coupling, the generator network and the inference network are learned jointly. Our method is applicable to semi-supervised learning in various data domains such as image, text, and tabular data. Our experiments demonstrate that our method performs well on semi-supervised learning tasks. 

\end{abstract}

\section{Introduction}

Generative modeling and likelihood-based learning is a principled framework for unsupervised and semi-supervised learning. The energy-based model (EBM) \cite{lecun2006tutorial}, being a generative version of a discriminator, is particularly convenient for semi-supervised learning. A notable recent paper is Grathwohl et al. \cite{grathwohl2019your}, which proposed the joint energy-based model (JEM) and applied it to classification. In an extension \cite{zhaojoint}, it was applied to semi-supervised learning. Earlier papers that touch on the relationship between modern EBM and classifier parametrized by deep networks \cite{lecunconvolutional, krizhevsky2012imagenet} include \cite{dai2014generative, xie2016theory, jin2017introspective, lazarow2017introspective}.

Recently, Pang et al. \cite{pang2020learning} proposed to learn EBM in latent space, where the EBM serves as a prior model for the latent vector of a generator model that maps the latent vector to the observed example. Both the EBM prior and the generator network can be learned jointly by maximum likelihood or its approximate or variational variants. The latent space EBM has been applied to image modeling, text modeling, and more recently molecule generation \cite{pang2020molecule}. See also the more recent developments \cite{aneja2020ncpvae, xiao2020vaebm}. 

Moving EBM from data space to latent space allows the EBM to stand on an already expressive generator model, and the EBM prior can be considered a correction of the non-informative uniform or isotropic Gaussian prior of the generator model. Due to the low dimensionality of the latent space, the EBM can be parametrized by a very small network, and yet it can capture regularities and rules in the data effectively (and implicitly). 

In this paper, we move the formulation of JEM \cite{grathwohl2019your} to latent space, so that the latent space EBM prior becomes an associative memory that couples the dense latent vector for generation and the one-hot symbolic vector for classification. Given the inferred latent vector, the conditional distribution of the one-hot vector is a regular softmax classifier based on the latent vector. We learn the symbol-vector coupling, the generator network, and the inference network jointly. Our method is applicable to semi-supervised learning in various data domains such as image, text, and tabular data. In experiments, our method demonstrates either competitive or improved performance compared to previous methods. 

The symbol-vector coupling in the latent space is of independent interest in terms of the interaction between symbolic reasoning and distributed representation. We also provide a general theoretical framework in appendix for learning latent EBM in general, where our learning method is a special case.

\section{Model and learning}

\subsection{Model: symbol-vector coupling}

Let $x \in \mathbb{R}^{D}$ be the observed example. Let $z \in \mathbb{R}^d$ be the latent vector. Let $y$ be the symbolic one-hot vector for classification of $K$ categories. $y$ is $K$-dimensional. Our model is defined by 
\begin{align}
    p_\theta(y, z, x) &= p_\alpha(y, z) p_\beta(x | z)
\end{align}
where $p_{\alpha}(y, z)$ is the prior model with parameters $\alpha$, $p_\beta(x|z)$ is the top-down generation model with parameters $\beta$, and $\theta = (\alpha, \beta)$. Given $z$, $y$ and $x$ are independent, i.e., $z$ is sufficient for $y$. After inferring $z$ from $x$, the inference of $y$ is based on $z$, i.e., $z$ is the information bottleneck \cite{tishby2015deep}. 

The prior model $p_{\alpha}(y, z)$ is formulated as an energy-based model,
\begin{align}
p_{\alpha}(y, z) = \frac{1}{Z_\alpha} \exp(\langle y, f_\alpha(z)\rangle) p_0(z), \label{eq:prior}
\end{align}   
where $p_0(z)$ is a reference distribution, assumed to be isotropic Gaussian (or uniform) prior of the conventional generator model. $f_{\alpha}(z) \in \mathbb{R}^K$ is parameterized by a small multi-layer perceptron. $Z_\alpha$ is the normalizing constant or partition function. 

The energy term $ \langle y, f_\alpha(z)\rangle$ in (\ref{eq:prior}) forms an associative memory that couples the symbol $y$ and the dense vector $z$. Given $z$, 
\begin{align} 
p_\alpha(y|z) \propto \exp(\langle y, f_\alpha(z)\rangle),  \label{eq:softmax}
\end{align}
 i.e., a softmax classifier, where $f_\alpha(z)$ provides the $K$ logit scores for the $K$ categories. Marginally, 
 \begin{align}
 p_\alpha(z) = \frac{1}{Z_\alpha}  \exp(F_\alpha(z))p_0(z),
 \end{align}
  where the marginal energy term
 \begin{align} 
 F_\alpha(z) = \log \sum_y \exp(\langle y, f_\alpha(z)\rangle), \label{eq:lse}
 \end{align} 
  i.e., the so-called log-sum-exponential form.  The summation can be easily computed because we only need to sum over $K$ different values of the one-hot $y$. 
  
The above prior model $p_\alpha(y, z)$ stands on a generation model $p_\beta(x|z)$, which is assumed to be the same as the top-down network in VAE \cite{kingma2013auto}. In particular, 
$x = g_\beta(z) + \epsilon$, 
where $\epsilon \sim {\rm N}(0, \sigma^2 I_D)$, so that $p_\beta(x|z) \sim {\rm N}(g_\beta(z), \sigma^2 I_D)$. As in VAE, $\sigma^2$ takes an assumed value. 

\subsection{Prior and posterior sampling: symbol-aware continuous vector computation} 

Sampling from the prior $p_\alpha(z)$ and the posterior $p_\theta(z|x)$ can be accomplished by Langevin dynamics. For prior sampling from $p_\alpha(z)$, Langevin dynamics iterates $z_{t+1} = z_t + s^2 \nabla_z \log p_\alpha(z_t)/2 + {\rm N}(0, s^2 I_d)$, where $s$ is the step size, and the gradient is computed by 
\begin{align} 
\nabla_z \log p_\alpha(z) = \E_{p_\alpha(y|z)}[\nabla_z \log p_\alpha(y, z)] =  \E_{p_\alpha(y|z)}[\langle y, \nabla_z f_\alpha(z)\rangle ] , 
\end{align} 
where the gradient computation involves averaging $\nabla_z f_\alpha(z)$ over the softmax classification $p_\alpha(y|z)$ in (\ref{eq:softmax}). Thus the sampling of the continuous dense vector $z$ is aware of the symbolic $y$. 

Posterior sampling from $p_\theta(z|x)$ follows a similar scheme, where 
\begin{align} 
\nabla_z \log p_\theta(z|x) = \E_{p_\alpha(y|z)}[\langle y, \nabla_z f_\alpha(z)\rangle ]  + \nabla_z \log p_\beta(x|z).
\end{align} 
When the dynamics is reasoning about $x$ by sampling the dense continuous vector $z$ from $p_\theta(z|x)$, it is aware of the symbolic $y$ via the softmax $p_\alpha(y|z)$. 

Thus $(y, z)$ forms a duality between symbol and dense vector, i.e., they are two sides of a single ``sym-vec coin.'' The entropy of $p_\alpha(y|z)$ may determine whether the symbolic interpretation will pop out from $z$. 

Pang et al. \cite{pang2020learning} proposed to use prior and posterior sampling for maximum likelihood learning. Due to the low-dimensionality of the latent space, MCMC sampling is affordable and mixes well. Comparing prior and posterior sampling, prior sampling is particularly affordable, because $f_\alpha(z)$ is a small network. In comparison, $\nabla_z \log p_\beta(x|z)$ in the posterior sampling requires back-propagation through the generator network, which can be more expensive. Therefore we shall amortize the posterior sampling from $p_\beta(x|z)$ by an inference network, as detailed in the next subsection. 

\subsection{Amortizing posterior sampling and variational learning}

Similar to VAE  \cite{kingma2013auto}, we recruit an inference network $q_\phi(z|x)$ to approximate the true posterior $p_\theta(z|x)$, in order to amortize posterior sampling. Following VAE, we learn the inference model $q_\phi(z|x)$ and the top-down model $p_\theta(z, x)$ jointly. 

For unlabeled $x$, the log-likelihood $\log p_\theta(x)$ is lower bounded by the evidence lower bound (ELBO),
\begin{align}
    \ELBO(\theta, \phi) &= \log p_\theta(x) - \KL(q_\phi(z|x)\|p_\theta(z|x)) \\
    &=  \E_{q_\phi(z|x)} \left[\log p_\beta(x|z)\right] - \KL(q_\phi(z|x) \| p_\alpha(z)), 
\end{align}
where $\KL$ denotes the Kullback-Leibler divergence. 

For the prior model, the learning gradient  is 
\begin{align}
    \nabla_\alpha \ELBO = \E_{q_\phi(z|x)} [\nabla_\alpha F_{\alpha}(z)] - \E_{p_\alpha(z)} [\nabla_\alpha F_{\alpha}(z)],
\end{align}
where $F_\alpha(z)$ is defined by (\ref{eq:lse}), $\E_{q_\phi(z|x)}$ is approximated by samples from the inference network, and $\E_{p_\alpha(z)}$ is approximated by persistent MCMC samples from the prior.

Let $\psi = \{\beta, \phi\}$ collect the parameters of the inference (encoder) and generator (decoder) models. The learning gradients for the two models are
\begin{align}
    \nabla_\psi \ELBO = \nabla_\psi \E_{q_\phi(z|x)} [\log p_\beta(x|z)] - \nabla_\psi \KL(q_\phi(z|x) \| p_0(z)) + \nabla_\psi \E_{q_{\phi(z|x)}} [F_{\alpha}(z)],
\end{align}
where $\KL(q_\phi(z|x) \| p_0(z))$ is tractable and the expectations in the other two terms are approximated by samples from the inference network $q_\phi(z|x)$ with reparametrization trick  \cite{kingma2013auto}. 

VAE is employed for training latent space EBM in recent work \cite{aneja2020ncpvae, xiao2020vaebm}. In our work, we train the inference network jointly with the latent space EBM. We only need to include the extra $F_\alpha(z)$ term, and $\log Z_\alpha$ is a constant that can be discarded. This expands the scope of VAE where the top-down model is a latent EBM. 

As mentioned above, we shall not amortize the prior sampling from $p_\alpha(z)$ due to its simplicity, and due to the fact that sampling $p_\alpha(z)$ is only needed in the training stage, but is not required in the testing stage. In general,  we can also amortize the sampling of $p_\alpha(z)$ by a flow-based model \cite{dinh2014nice, dinh2016density}. See appendix for a general theoretical formulation. 

\subsection{Labeled data} 

For a labeled example $(x, y)$, the log-likelihood can be decomposed into $\log p_\theta(x, y) = \log p_\theta(x) + \log p_\theta(y |x)$. The gradient of $\log p_\theta(x)$ can be computed in the same way as the unlabeled data, and 
\begin{align} 
    p_\theta(y|x) = \E_{p_\theta(z|x)} [p_\theta(y|z)] \approx \E_{q_\phi(z|x)} [p_\theta(y|z)], 
 \end{align} 
 where $p_\theta(y|z)$ is the softmax classifier defined by (\ref{eq:softmax}), and $q_\phi(z|x)$ is the learned inference network. Thus $\nabla_\theta \log p_\theta(y|x) \approx \nabla_\theta \log  \E_{q_\phi(z|x)} [p_\theta(y|z)]$. The gradients backpropagate to both the EBM prior and the inference network.
 
 For semi-supervised learning, we can combine the learning gradients from both unlabeled and labeled data. 

\subsection{Algorithm}

The learning and sampling algorithm is described in Algorithm \ref{algo:short}. 

\begin{algorithm}[H]
	\SetKwInOut{Input}{input} \SetKwInOut{Output}{output}
	\DontPrintSemicolon
	\Input{Learning iterations~$T$, learning rates~$(\eta_0,\eta_1,\eta_2)$, initial parameters~$(\alpha_0, \beta_0, \phi_0)$, observed unlabelled examples~$\{x_i\}_{i=1}^M$, observed labelled examples~$\{(x_i, y_i)\}_{i=M+1}^{M+N}$, unlabelled and labelled batch sizes $(n,m)$, set of persistent chains~$\{z_i^{-}\sim p_0(z)\}_{i=1}^{L}$, and number of Langevin dynamics steps $T_{LD}$.}
	\Output{ $(\alpha_{T}, \beta_{T}, \phi_{T})$.}
	\For{$t = 0:T-1$}{			
		\smallskip
		1. {\bf Mini-batch}: Sample unlabelled $\{ x_i \}_{i=1}^m$ and labelled observed examples $\{ x_i, y_i \}_{i=m+1}^{m+n}$. \\
		2. {\bf Prior sampling}: For each unlabelled $x_i$, randomly pick and update a persistent chain $z_i^{-}$ by MCMC with target distribution $p_\alpha(z)$ for $T_{LD}$ steps. \\
		3. {\bf Posterior sampling}: For each $x_i$, sample $z_i^{+} \sim q_\phi(z|x_i)$ using the inference network and reparameterization trick. \\
		4. {\bf Unsupervised learning of prior model}:\\
		$\alpha_{t+1} = \alpha_t + \eta_0 \frac{1}{m}\sum_{i=1}^{m} [\nabla_\alpha F_{\alpha_t}(z_i^{+}) - \nabla_\alpha F_{\alpha_t}(z_i^{-})]$. \\
		5. {\bf Unsupervised learning of inference and generator models}:\\ $\psi_{t+1} = \psi_t + \eta_1 \frac{1}{m}\sum_{i=1}^{m}[\nabla_\psi [\log p_{\beta_t}(x|z_i^{+})] - \nabla_\psi \KL(q_{\phi_t}(z|x_i) \| p_0(z)) + \nabla_\psi[F_{\alpha_t}(z_i^{+})]$. \\
		6. {\bf Supervised learning of prior and inference model}:\\
		$\theta_{t+1} = \theta_t + \eta_2 \frac{1}{n}\sum_{i=m+1}^{m+n} \sum_{k=1}^K y_{i,k} \log (p_{\theta_t}(y_{i,k}|z_i^{+}))$.
	}
	\caption{Semi-supervised learning of latent space EBM with symbol-vector coupling.}
	\label{algo:short}
\end{algorithm}


\section{Related work} 

\textbf{Discriminative models.} Recent semi-supervised learning (SSL) methods based on discriminative models have achieved substantial progress. Most of these methods rely on data augmentation strategies which heavily exploit class-invariance properties of images \cite{berthelot2019mixmatch, sohn2020fixmatch}. Invariance properties for text and tabular data are however less clear. One successful method, virtual adversarial training (VAT) \cite{miyato2018virtual}, does not depend on image-domain-specific properties. Instead, it finds an adversarial augmentation to $x$ within an $\epsilon$-ball of $x$ with respect to $l_p$ norm such that the distance between the class distribution conditional on $x$ and that conditional on the augmentation is maximized. This approach is applicable to continuous data beyond images like tabular data. It nevertheless still has the limitation on discrete data since the adversarial augmentation requires the data space to be differentiable.

\textbf{GANs.} Generative models in the GAN family have also been applied to SSL and exhibited promising results. Salimans et al. \cite{salimans2016improved} changed the discriminator in GAN to be a classifier with an extra "generated" class, which is thus able to simultaneously distinguish real and generated samples and to predict class labels. To avoid the incompatible roles of the classifier and discriminator, Li et al. \cite{chongxuan2017triple} proposed to keep the discriminator in the original GAN and add a classifier for class prediction. Dai et al. \cite{dai2017good} argued that a preferred generator for SSL should be a complement generator in the sense that it should assign high densities for data points with low densities in the data distribution. These GAN-based SSL methods generally perform well on images, competitive with discriminative models. However, GANs are highly tuned to image data. The adversarial generative training relies on the differentibility of data generation and applying it to discrete data like text with does not exhibit much gain if no harm is caused \cite{caccia2019language}.

\textbf{Energy-based model.} Recently EBMs have also been applied to SSL. Grathwohl et al. \cite{grathwohl2019your} proposed Joint Energy-based Model (JEM) on top of a regular discriminative classifier by re-interpreting the logits. Zhao et al. \cite{zhaojoint} extended JEM to SSL and obtained reasonable performance. With a similar formulation, Gao et al. \cite{gao2020flow} adversarially trained the EBM with a pretrained GLOW \cite{kingma2018glow} and obtained performance competitive with GAN-based and discriminative models, especially after it is combined with VAT. Our model shares similarities with these models in that our model has a latent space EBM and it is constructed through re-using the logits of a discriminative classifier in the latent space. Our EBM in the latent space however has a much lower dimension and smoother landscape, which renders fast sampling and mixing. The learning also does not require expensive pretraining of GLOW like in \cite{gao2020flow}. Also, it is not easy to apply data space EBM to discrete data since its training requires either the data space or the generation process of the auxiliary generator (e.g., GLOW) to be differentiable.

\textbf{Likelihood-based models.} In a seminal work \cite{kingma2014semi}, Kingma et al. adopted VAE for SSL where a discrete latent variable, $y$, representing the class labels is introduced in addition to the usual continuous vector, $z$. In \cite{makhzani2015adversarial}, adversarial training was applied to force $y$ and $z$ to capture disentangled semantics. In our model, the continuous vector is coupled with the discrete variable or symbol to form a latent space EBM prior. More recently, Izmailov et al. \cite{izmailov2019semi} proposed a flow-based model, FlowGMM, where the data distribution is mapped to a Gaussian mixture in the latent space with invertible flow. Our model is similar in the sense that it uses an inference network to map the data distribution to an latent space EBM. But our model does not have the restriction of invertibility. FlowGMM is aimed for broad applicability and demonstrate promising results on text and tabular data, sharing the goal of our work.

\section{Experiments}
We evaluate our model on a wide variety of datasets to demonstrate the broad applicability of our model. We begin the evaluation with image data which most SSL research works have been focused on. Next we consider text data in both continuous and discrete situations. Finally, our model is assessed on tabular data. Two recent works applied their methods, FlowGMM \cite{izmailov2019semi} and JEM \cite{zhaojoint}, beyond image data and these two models are closely related to ours. They thus serve as important baselines of our model. 

For image experiments, the inference network is implemented with a standard Wide ResNet \cite{zagoruyko2016wide}, widely used in image SSL works \cite{oliver2018realistic}. The generator network is a 4-layer deconvoluational network, similar to the generator in DCGAN \cite{radford2015unsupervised}. The inference and generator networks in text and tabular experiments are all a 3-layer MLP. The latent EBM prior in all experiments is also a small 3-layer MLP with a hidden dimension of 200. All model parameters are initialized with Xavier normal \cite{glorot2010understanding}. Adam \cite{kingma2015adam} is adopted for all model optimization. The models are trained until convergence (taking approximately 400,000 parameter updates for image models, 100,000 updates for text models without pretrained embeddings, 3,000 updates for text models with pretrained embeddings, 4,000 updates for tabular models). We run 20 steps of persistent chains with step size $0.6$ to obtain MCMC samples used in learning the latent space EBM prior (i.e., Step 4 in Algorithm~\ref{algo:short}). Notice that MCMC sampling only needs to compute the gradients of the small latent space EBM and does not need to backpropagate through the larger generator.  

\subsection{Image}
Standard image SSL benchmarks, SVHN \cite{netzer2011reading}, with 1,000 labeled data and 64,932 unlabeled data, and CIFAR-10 \cite{cifar10}, with 4,000 labeled data and 41,000 unlabeled data, are adopted to assess our model on image classification. The results are summarized in Tabel~\ref{tab:image}. Besides FlowGMM \cite{izmailov2019semi} and JEM \cite{zhaojoint}, we also compare our model to other likelihood-based models, VAE M1+M2 \cite{kingma2014semi}, AAE \cite{makhzani2015adversarial}, in addition to representative GAN-based models, TripleGAN \cite{chongxuan2017triple}, BadGAN \cite{dai2017good} and discriminative models, $\Pi$-Model \cite{laine2016temporal} and VAT \cite{miyato2018virtual}. Our model outperforms FlowGMM and JEM and other likelihood-based models. The improvement is especially clear on SVHN (with almost $10\%$ absolute improvement compared to FlowGMM). We however observe a performance gap between our model and GAN-based and discriminative methods which highly tuned for images.

\begin{table}[h]
    \centering
    \vskip 0.1in
    \begin{tabular}{lcc}
      \toprule
       & SVHN & CIFAR-10 \\
      Method & 1000 Labels & 4000 Labels \\
      \midrule
      VAE M1+M2 & 64.0 $\pm$ 0.1 & - \\
      AAE & 82.3 $\pm$ 0.3 & - \\
      JEM & 66.0 $\pm$ 0.7  & - \\
      FlowGMM  & 82.4 & 78.2 \\
      \textbf{Ours}  & 92.0 $\pm$ 0.1  & 78.6 $\pm$ 0.3 \\
      \midrule
      TripleGAN & 94.2 $\pm$ 0.2 & 83.0 $\pm$ 0.4 \\
      BadGAN  & 95.8 $\pm$ 0.03  & 85.6 $\pm$ 0.03  \\
      $\Pi$-Model & 94.6 $\pm$ 0.2  & 83.6 $\pm$ 0.3 \\
      VAT & 96.3 $\pm$ 0.1 & 88.0 $\pm$ 0.1 \\
      \bottomrule
    \end{tabular}
    \caption{Accuracy on SVHN and CIFAR-10.}
    \label{tab:image}
  \vskip -0.2in
\end{table}

\subsection{Text}
We evaluate our model on text with a widely used text classification dataset, AGNews \cite{zhang2015character},  in two settings. The first one follows the evaluation protocol used in FlowGMM \cite{izmailov2019semi}. Text data are first embedded by a pretrained language model, BERT \cite{devlin2019bert}, and the SSL model is constructed on the embedding space. We use the preprocesed data provided by \cite{izmailov2019semi} which contains 200 labeled data and 120,000 unlabled data are used in training. AGNews has 4 classes. Our model is compared to FlowGMM \cite{izmailov2019semi}, RBF Label Spreading (a graph-based label spreading method) \cite{zhou2004learning}, and $\Pi$-Model \cite{laine2016temporal}. Table~\ref{tab:text-bert} displays the results. Our model exhibits competitive performance with FlowGMM and outperforms other baselines.

The second setting where a large pretrained language model is not available follows the setup used in Gururangan et al. \cite{gururangan2019variational} where they proposed VAMPIRE. While the huge amount of text data required for large transformer-based language model pretraining is available for some languages, such as English, this scale of data is not available for all languages. We use the preprocessed data provided by \cite{gururangan2019variational} and it contains 200 labeled data and 114,600 unlabeled documents for training. Follow \cite{gururangan2019variational}, a document is modeled by the unigram of its words. Thus, each document is a vector of vocabulary size, $V$ ($V=30,000$ for AGNews), and each element represents a word's occurring frequency in the document, modeled by a multinominal distribution. Notice that most popular SSL methods, such as GAN-based models, data space EBM, and VAT, cannot be easily applied in this case since the data space is discrete. We compare our model to VAMPIRE using unigram representation, self-training, and supervised training with Glove word embeddings pretrained on in-domain and out-domain data (see \cite{gururangan2019variational} for detailed descriptions of these baselines). The results are summarized in Table~\ref{tab:text-bow}. Our model clearly outperforms these baselines. It is worth pointing that FlowGMM is also applicable in this setting. We attempted to apply FlowGMM to this task but achieved low accuracy.

\begin{table}[h]
    \centering
    \vskip 0.1in
    \begin{tabular}{lcc}
      \toprule
       & AGNews-Bert\\
      Method & 200 Labels \\
      \midrule
      RBF Label Spreading & 36.1\\
      FlowGMM  & 82.1 $\pm$ 1.0 \\
      \textbf{Ours}  & 82.0 $\pm$ 0.2 \\
      \midrule
      $\Pi$-Model & 80.2 $\pm$ 0.3 \\
      \bottomrule
    \end{tabular}
    \caption{Accuracy on AGNews with Bert embeddings.}
    \label{tab:text-bert}
  \vskip -0.2in
\end{table}

\begin{table}[h]
    \centering
    \vskip 0.1in
    \begin{tabular}{lcc}
      \toprule
       & AGNews-Unigram\\
      Method & 200 Labels \\
      \midrule
      Self-training & 77.3 $\pm$ 1.7\\
      Glove (ID) & 70.4 $\pm$ 1.2\\
      Glove (OD)  & 68.8 $\pm$ 5.7\\
      VAMPIRE  & 81.9 $\pm$ 0.5\\
      \textbf{Ours}  & 84.5 $\pm$ 0.3 \\
      \bottomrule
    \end{tabular}
    \caption{Accuracy on AGNews with Unigram.}
     \label{tab:text-bow}
  \vskip -0.2in
\end{table}

\subsection{Tabular data}
We use three tabular datasets from the UCI repository. Hepmass and Miniboone were utilized in \cite{izmailov2019semi} for SSL and Protein was used in \cite{zhaojoint}. Protein has a continuous target variable and we follow \cite{zhaojoint} to bin the targets into 10 equally weighted buckets. We use the same experimental settings as in the two prior works. The number of labeled / unlabeled data for Hepmass, Miniboone, and Protein are 20/140,000, 20/65,000, 100/41,057 respectively. Hepmass and Miniboone have 2 classes, while Protein has 10 classes.   We compare our model to RBF Label Spreading, JEM, FlowGMM, $\Pi$-Model, and VAT. Tabel~\ref{tab:tabular} summarizes the results. Our model outperforms all baselines across the three tabular datasets.

\begin{table}[h]
    \centering
    \vskip 0.1in
    \begin{tabular}{lccc}
      \toprule
       & Hepmass & Miniboone & Protein\\
      Method & 20 Labels & 20 Labels & 100 Labels\\
      \midrule
      RBF Label Spreading & 84.9 & 79.3 & -\\
      JEM & - & - & 19.6 \\
      FlowGMM  & 88.5 $\pm$ 0.2 & 80.5 $\pm$ 0.7 & -\\
      \textbf{Ours}  & 89.1 $\pm$ 0.1 & 81.2 $\pm$ 0.3 & 23.1 $\pm$ 0.3\\
      \midrule
      $\Pi$-Model & 87.9 $\pm$ 0.2 & 80.8 $\pm$ 0.01 & -\\
      VAT & - & - & 17.1\\
      \bottomrule
    \end{tabular}
    \caption{Accuracy on Hepmass, Miniboone, and Protein.}
    \label{tab:tabular}
  \vskip -0.2in
\end{table}

\section{Conclusion}

Semi-supervised learning based on latent space EBM prior with symbol-vector coupling is very natural. For unlabeled data, the marginal EBM prior is in the form of sum of exponentials. For labeled data, the conditional distribution of label given the inferred latent vector is a regular softmax classifier. The semi-supervised learning can be based on a principled likelihood-based framework, with inference computation being amortized by a variational inference network. 

Our model may be interpreted as a generative classifier, where the latent vector used for classification is inferred based on a top-down generative model. The top-down model and the posterior inference captures the concept of information bottleneck \cite{tishby2015deep} more naturally than bottom-up classifier. The posterior inference of a top-down model may be more robust to adversarial perturbations than a classifier defined on the input directly, because the posterior inference can explain away the adversarial perturbations via the top-down model. The inference of the latent vector is aware of the underlying symbol, and the symbol-vector coupling in our prior model may shed light on the interaction between symbolic reasoning and continuous computation. We may consider a multi-layer top-down model where each layer consists of dense sub-vectors coupled with symbolic one-hot sub-vectors, so that continuous computation based on dense sub-vectors is aware of the corresponding symbols. 

Our experiments show that our semi-supervised learning method outperforms existing methods on text and tabular data. We shall continue to improve our method on image data. To rephrase the title of this workshop, we cannot believe our principled model-based semi-supervised learning method does not work better than existing methods on image data. 

\section*{Appendix: learning latent EBM with divergence perturbation formulation} 

This section provides a theoretical formulation for learning latent EBM in general, where our model is a special case. We can embed our learning method within this general formulation. 

The latent EBM is of the following form: 
\begin{align} 
  p_\theta(z, x) = \frac{1}{Z_\theta} \exp(f_\theta(x, z)). 
\end{align} 
Marginally, $p_\theta(x) = \int p_\theta(z, x) dz$, and the posterior is $p_\theta(z|x) = p_\theta(z, x)/p_\theta(x)$. 

Let $p_{\rm data}(x)$ be the data distribution that generates $x$. Maximum likelihood estimation (MLE) minimizes $\KL(p_{\rm data}(x) \| p_\theta(x))$, where expectation with respect to $p_{\rm data}$ can be approximated by averaging over observed examples. 

Learning of latent EBM can be based on the following divergence perturbation (see \cite{han2019divergence, han2020joint} for related formulations), 
\begin{align} 
  \Delta =  \KL(p_{\rm data}(x) \| p_\theta(x))  + \KL(q_{+}(z|x) \| p_\theta(z|x)) - \KL(q_{-}(z, x) \| p_\theta(z, x)), \label{eq:tri}
\end{align}
which is a perturbation of the divergence $\KL(p_{\rm data}(x) \| p_\theta(x))$ that underlies MLE. $q_{+}(z|x)$ is the positive phase sampler, and $q_{-}(z, x)$ is the negative phase sampler. They correspond to the positive phase and negative phase of MLE learning of latent EBM such as Boltzmann machine \cite{hinton1985boltzmann}. For two distributions $q(z|x)$ and $p(z|x)$, we define $\KL(q(z|x)\|p(z|x)) = \E_{p_{\rm data}(x)} \E_{q(z|x)} [ \log (q(z|x)/p(z|x)]$, with an outer expectation with respect to $p_{\rm data}$. The divergence $\KL(p_{\rm data}(x) \| p_\theta(x))$ for MLE is perturbed by the positive phase divergence $ \KL(q_{+}(z|x) \| p_\theta(z|x)) $ and the negative phase divergence $\KL(q_{-}(z, x) \| p_\theta(z, x))$. 

For latent space EBM prior, we have the factorization $p_\theta(z, x) = p_\alpha(z) p_\beta(x|z)$, and $q_{-}(z, x) = q_{-}(z) p_\beta(x|z)$, so that the negative phase $\KL$ in (\ref{eq:tri})  becomes $\KL(q_{-}(z) \| p_\alpha(z))$, i.e., the KL divergence between the priors. 

The above formulation has the following features. 

(1) Gradient of divergence perturbation $\Delta$ can be calculated explicitly without intractable integrations in $p_\theta(x)$ and $\log Z_\theta$, because $p_\theta(x)$ is merged with $p_\theta(z|x)$ to become $p_\theta(z, x)$ due to the positive phase $\KL$ term, and the $\log Z_\theta$ is canceled by the negative phase $\KL$ term. 

(2) For MLE learning, at iteration $t$, let $\theta_t$ be the current estimate of $\theta$. We let $q_{+}(z|x) = p_{\theta_t}(z|x)$ and $q_{-}(z, x)  = p_{\theta_t}(z, x)$. In this case, the positive phase $\KL(q_{+}(z|x) \| p_\theta(z|x))$ achieves its minimum 0 at $\theta_t$, so that its gradient at $\theta_t$ is 0. Similarly the negative phase $\KL(q_{-}(z, x) \| p_\theta(z, x))$ achieves its minimum 0 at $\theta_t$ too, so that its gradient at $\theta_t$ is also 0. Thus the gradient of the divergence perturbation $\Delta$ in (\ref{eq:tri}) is the same as the MLE gradient of the first KL term $ \KL(p_{\rm data}(x) \| p_\theta(x)) $. 

(3) For amortized computation, $q_{+}(z|x) = q_{\phi_{+}}(z|x)$ is an inference network, and $q_{-}(z, x) = q_{\phi_{-}}(z, x)$ is a synthesis network, with parameters $\phi_{+}$ and $\phi_{-}$ respectively, that are independent of the model parameter $\theta$. The learning can be based on $\min_\theta \min_{\phi_{+}} \max_{\phi_{-}} \Delta$. The positive phase $\KL$ leads to variational learning. The negative phase $\KL$ leads to a generalized form of contrastive divergence \cite{hinton2002cd}, as well as adversarial interpretation due to the negative sign and $\max_{\phi_{-}}$. 

For latent space EBM prior, we only need a synthesis network $q_{\phi_{-}}(z)$ to approximate the prior model $p_\alpha(z)$. We may consider using a flow-based model \cite{dinh2014nice, dinh2016density} for $q_{\phi_{-}}(z)$. 

In our current work, we employ an inference network for $q_{\phi_{+}}(z|x)$, but we leave $q_{-}(z)$ to MCMC sampling, because $p_\alpha(z)$ is very simple, and sampling of $p_\alpha(z)$ does not require back-propagation through the generator network. Moreover, sampling $p_\alpha(z)$ is not needed in the testing stage. 

(4) In between (2) and (3), both $q_{+}(z|x)$ and $q_{-}(z, x)$ can be sampled by short-run MCMC for inference and synthesis \cite{nijkamp2019learning, nijkamp2020learning}. The short-run MCMC runs a fixed number of MCMC iterations from a fixed initial distribution, so that both $q_{+}(z|x)$ and $q_{-}(z, x)$ are well defined. The gradient of $\Delta$ in (\ref{eq:tri}) based on short-run MCMC is a perturbation of the MLE learning gradient. The MLE learning gradient is impractical because it requires convergence of MCMC. Learning algorithm based on short-run MCMC converges to the solution of an estimating equation which is a perturbation of the MLE estimating equation. The learning algorithm of Pang et al. \cite{pang2020learning} works within this scheme. 

The above formulation encompasses many generative models and associated learning algorithms. It helps us understand the model and learning algorithm proposed in this paper. It will also help us develop new models and algorithms. 

\bibliographystyle{ieee_fullname}
\bibliography{main}

\end{document}